\documentclass{article}

\usepackage[square,sort,comma,numbers]{natbib}
\usepackage{nips_2017}
\usepackage{wrapfig,lipsum,booktabs}

\usepackage[utf8]{inputenc} 
\usepackage[T1]{fontenc}    
\usepackage{hyperref}       
\usepackage{url}            
\usepackage{booktabs}       
\usepackage{amsfonts}       
\usepackage{nicefrac}       
\usepackage{microtype}      

\usepackage{multicol}
\usepackage{changes}
\usepackage{booktabs}
\usepackage{graphicx}
\usepackage{subfig}
\usepackage{xparse}

\usepackage{cancel}

\RequirePackage{amsthm,amsmath}
\usepackage{graphicx, amssymb, mathrsfs,amsfonts,url, bm}

\usepackage{amsfonts}
\usepackage{amsfonts,amsthm,bm} 
\usepackage[boxed]{algorithm2e}

\DeclareMathOperator*{\argmin}{argmin}

\newcommand*{\argminl}{\argmin\limits}
\newcommand{\bsum}{\sum\limits}
\newcommand{\bW}{\mathbf{W}}
\newcommand{\ba}{\mathbf{A}}

\newcommand{\bo}{\mathbf{O}}

\newcommand{\bX}{\mathbf{X}}
\newcommand{\bR}{\mathbf{R}}

\newcommand{\bc}{\mathbf{C}}
\newcommand{\bbeta}{\bm{\beta}}
\newtheorem{theorem}{Theorem}[section]

\newtheorem{lemma}[theorem]{Lemma}

\usepackage{algcompatible,lipsum}
\algnewcommand{\lst}{\texttt{lst}}
\algnewcommand{\slst}{\texttt{slst}}
\algnewcommand{\SEND}{\textbf{send}}

\newsavebox{\algleft}
\newsavebox{\algright}

\title{$CompNet$: Neural networks growing via the compact network morphism}

\author{
   Jun Lu\thanks{Equal contribution. } , \,\,\,Wei Ma$^*$, \,\,\,Boi Faltings \\
   Swiss Federal Institute of Technology (EPFL), Lausanne, Switzerland  \\
   \texttt{junlulocky@gmail.com}, \texttt{ma.wei@epfl.ch}, \texttt{boi.faltings@epfl.ch} \\   
}

\begin{document}

\maketitle

\begin{abstract}
It is often the case that the performance of a neural network can be improved by adding layers.
In real-world practices, we always train dozens of neural network architectures in parallel which is a wasteful process. We explored $CompNet$, in which case we morph a well-trained neural network to a deeper one where network function can be
preserved and the added layer is compact.
The work of the paper makes two contributions: a). The modified network can converge fast and keep the same functionality so that we do not need to train from scratch again; b). The layer size of the added layer in the neural network is controlled by removing the redundant parameters with sparse optimization. This differs
from previous network morphism approaches which tend to add more neurons or channels beyond the actual requirements and result in redundance of the model. 
The method is illustrated using several neural network structures on different data sets including MNIST and CIFAR10.
\end{abstract}
\section{Introduction} \label{Introduction}
Over the recent decades, deep learning and neural networks have been used in a wide range of applications, e.g. computer vision \cite{rowley1998neural,ren2015faster,krizhevsky2012imagenet,ciregan2012multi}, 
natural language processing \cite{bahdanau2014neural,cho2014properties,kalchbrenner2013recurrent}, 
financial forecasting and time series
\cite{kim2006artificial,frank2001time,lu2017machine}. All of these applications need to train the neural networks in several days or even more than a month. If it turns out that we need to modify the neural network structure, it is either an immense waste of computational resources, or extremely prolonged experimentation cycles to train from scratch again.
This kind of modification to the well trained neural network structure can usually happen in tasks where the amount of data increases, where we may always try to add additional layers to the old structure.

In the meanwhile, deep learning requires huge computing resources and memory. In practice, a device designed for training the network may not have enough resource or time to train a complex neural network, e.g., mobile devices usually have small storage and limited computing ability. We also often make modifications to the neural network according to the training and validation performance and then retrain the new model. The process is time consuming and wasteful. Our work accelerates retrain the process and makes the new network model keep the same function as the previous one. 

In order to accelerate the training and exploration of deep neural network structure, there has been already some works on making a new network inherit the knowledge of the parent model for the same task. Net2Net \cite{chen2015net2net} proposes a method based on the concept of function-preserving transformation. But Net2Net  can only add a layer with specific neuron size. Network morphism (NetMorph) \cite{wei2016network} approaches the inheritance problem by deconvolutional operation and actually there is no criteria to select the neuron size and it always makes the network structure much larger than expected. Specifically speaking, Net2Net and NetMorph only consider increasing the size of the neural network but not limiting the growth of complexity.

On ther other hand, researchers tend to modify the neural network strucutre empirically. Although there are some works on constructing the neural network automatically, e.g., Google \cite{DBLP:journals/corr/abs-1709-07417,DBLP:journals/corr/abs-1710-05941} has been exploring ways to automate the design of machine learning models and \cite{BurnaevKNK16} proposes an approach for automatic construction of binary classifier of Recurrent Neural Network (RNN). However, most of the researchers look for a good architecture manually and empirically. In the paper, we propose a framework to speed up and control the process of increasing the model size.

Our method is illustrated on multilayer perceptron (MLP) and convolutional network. \cite{ma2017equivalence} shows that a convolutional layer can be transformed to a dense layer, which does not affect the learning process. Based on the conclusion, the traditional methods that work over matrix multiplication can be applied to the convolutional layers easily.

Recently, some compression methods are used to reduce the model size in deep learning. \cite{abs-1710-09282} investigates into the methods of compression of neural network and summarizes four types of methods: a). Parameter pruning and sharing. b). Low-rank factorization. c). Transfer/compact convolutional filters. d). Knowledge distillation \cite{HintonVD15}. 
For feature selection, \cite{muthukrishnan2016lasso} compares Lasso, Ordinary Least Square and ridge regression and finds that Lasso can work better than the others. \cite{DBLP:journals/corr/HeZS17} uses Lasso to accelerate very deep convolutional neural networks.  \cite{2017arXiv171101796T} points out that the traditional Lasso tends to keep the relevant features and some independent but important features sometimes are removed. They solve the problem by penalizing the similarity matrix of the features. In our framework, we adopt this variant of Lasso. 


\textbf{Notation}: In the paper, scalar variables are written as non-bold font lowercases, e.g., $c$ and $s$ are scalar values. Matrices and vectors are written as bold font capitals or bold font lowercases. For example, $\mathbf{W} \in \mathbb{R}^{a \times b}$ represents a matrix of size $a \times b$. We use $\mathbf{W}_i$ to denote the $i^{th}$ column of matrix $\mathbf{W}$, $\mathbf{W}_{j,:}$ to denote the $j^{th}$ row of matrix $\mathbf{W}$. We use the superscripts to indicate the layer index of the neural network, e.g. $L^{(1)}$ is the first layer of the structure.



\section{Proposed Method} \label{methodology}
When operating in a continuous learning setting such as reinforcement learning, we tend to use deeper neural networks to get better result without overfitting as we collect more data. And training the new neural network from scratch can be extremely time consuming.
In this section, we describe our proposed algorithms. Briefly speaking, our methods take two steps. In the first step, we generate the child model by adding an additional layer and make it inherit the ability from the parent model by regression algorithms. 
In the second step, we use sparse optimization methods (e.g. Lasso) to reduce the size of the new layers. To illustrate, we use multilayer perceptron (MLP) as an example as shown Figure~\ref{mlpnet}. But our work can be extended to convolutional neural networks (CNN) easily from the equivalence between fully connected layers and convolutional layers \cite{gal2016uncertainty,ma2017equivalence}. 

\subsection{Problem and Notation}\label{sec:problem}
\begin{figure}[!ht]
	\centering
	\subfloat[MLP before inserting one layer]{%
		\includegraphics[height=0.275\textwidth]{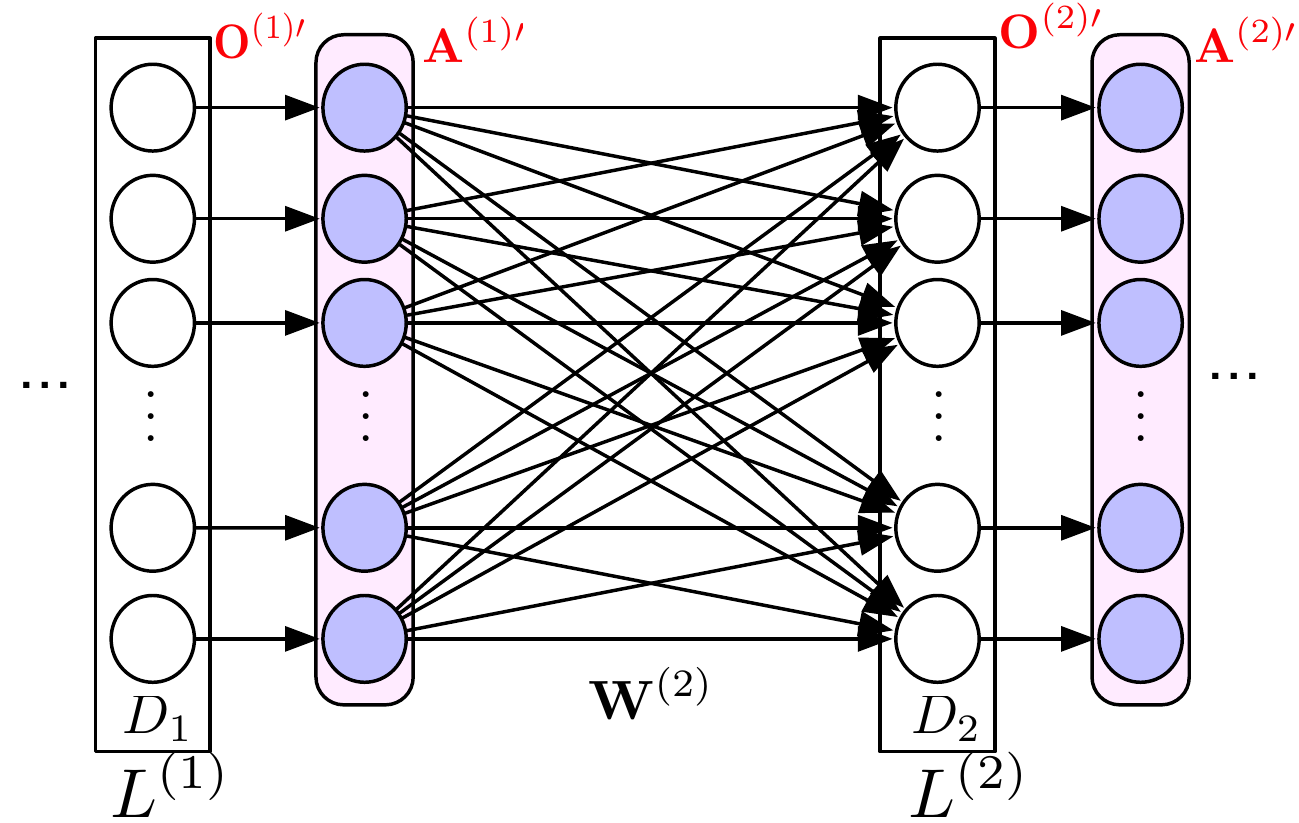}
		\label{mlp1}
	} 
	~
	\subfloat[MLP after inserting one layer]{%
		\includegraphics[height=0.275\textwidth]{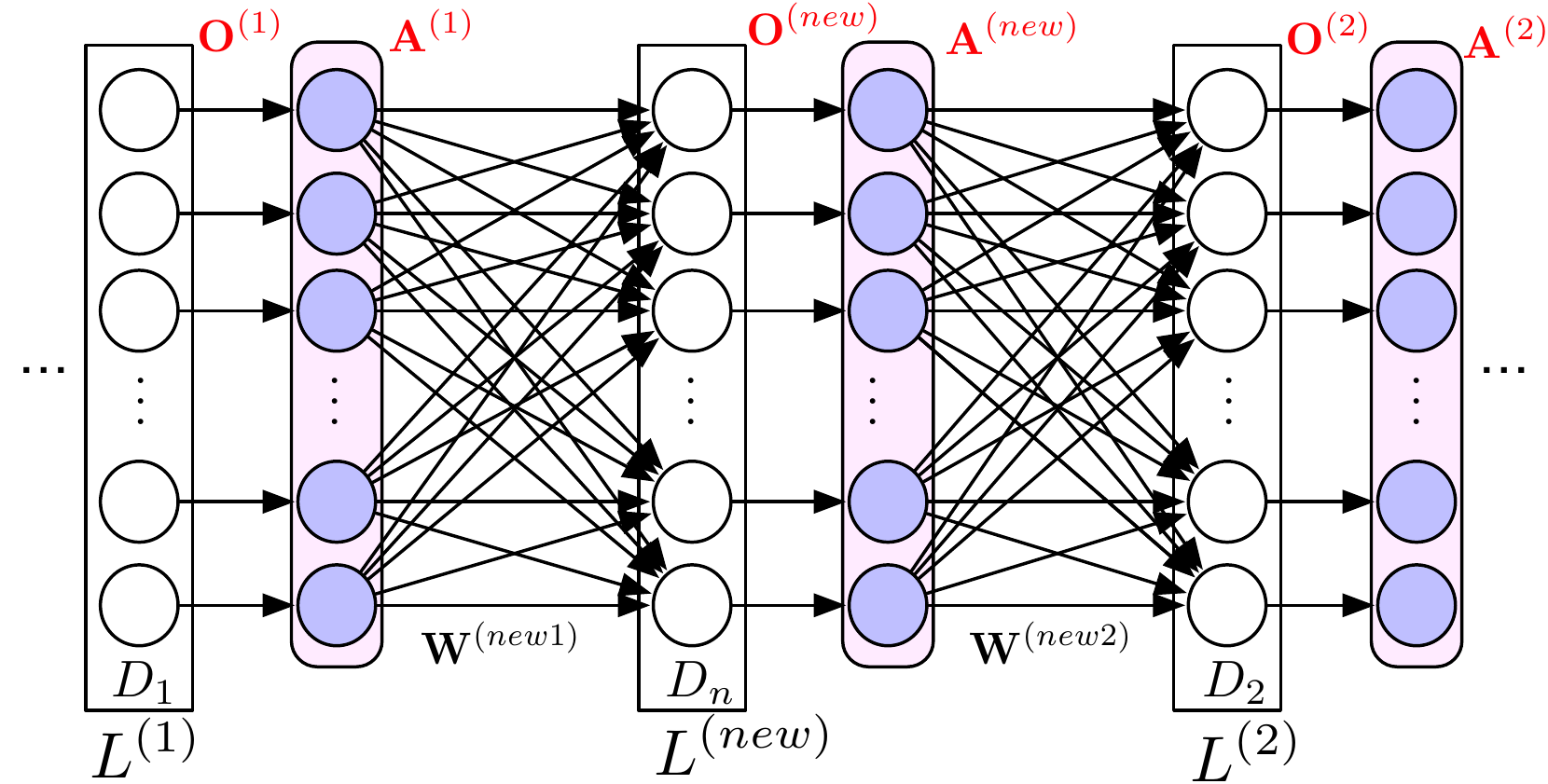}
		\label{mlp2}
	}
	\caption{\textbf{Notations for our approach}. Figure~\ref{mlp1} is a part of the parent network and Figure~\ref{mlp2} is a child network after inserting a new layer into the part of Figure~\ref{mlp1}. $D_{d}$ is the number of the neurons in the layers and $L^{(d)}$ is the layer name, where $d \in \{1,2,n\}$. $\bW^{(2)}$ and $\bW^{(new2)}$ are the weights of $L^{(2)}$ in Figure~\ref{mlp1} and Figure~\ref{mlp2} respectively. $\bW^{(new1)}$ is the weights of the new layer $L^{(new)}$. The shallow pink box is the activation function layer. Note here $\bo^{(1)} = \bo^{(1)\prime}$ and $\ba^{(1)} = \ba^{(1)\prime}$.}
	\label{mlpnet}
\end{figure}

Consider the fully connected layer structure shown in Figure~\ref{mlpnet}.
Figure~\ref{mlp1} is a part of a well-tuned parent network and it represents two-layer structure of the parent network. Figure~\ref{mlp2} is a child network after inserting a fully connected layer into the part of Figure~\ref{mlp1}. We denote the output of layer $L^{(1)}$ and $L^{(2)}$ in Figure~\ref{mlp1} by $\bo^{(1)\prime}$ and $\bo^{(2)\prime}$ respectively. And we denote the activation output of layer $L^{(1)}$ and $L^{(2)}$ in Figure~\ref{mlp1} by $\ba^{(1)\prime}$ and $\ba^{(2)\prime}$ respectively. Note here that we use a prime to indicate whether it is the output or activation output of each layer in the original setting or not.
In Figure~\ref{mlp2}, we denote the output of layers $L^{(1)}$, $L^{(new)}$ and $L^{(2)}$  to be $\bo^{(1)}$, $\bo^{(new)}$ and $\bo^{(2)}$ respectively. $\ba^{(1)}$, $\ba^{(new)}$ and $\ba^{(2)}$ are the activation output of the corresponding layers. The activation function is denoted by $h(x)$, i.e. $\ba^{d} = h(\bo^{d}), \forall d \in \{(1), (2), (new), (1)\prime, (2)\prime\}$. 



As stated, we have two targets: a). let the child network inherit the competence and preserve the functionality of the parent network. Thus after morphing, we want $\ba^{(2)}$ to be as close to $\ba^{(2)\prime}$ as possible, thus $\bo^{(2)}$ is as close to $\bo^{(2)\prime}$ as possible; b). sparsify the new layer to control the increase of complexity and reduce correlated neurons, i.e. we want the size of layer $L^{(new)}$ to be compact. 


\subsection{Algorithm}
We propose three algorithms to add an additional layer into the well trained neural network compactly. Algorithm~\ref{algframe} and~\ref{altnativeupdating} reduce the neuron size of $L^{(new)}$ directly by applying the sparse optimization to $\mathbf{W}^{(new1)}$. And Algorithm~\ref{pruningbyw2} reduces the neuron size of $L^{(new)}$ by applying the sparse optimization to $\mathbf{W}^{(new2)}$.

\subsubsection{Reduce the neuron size of $L^{(new)}$ from $\mathbf{W}^{(new1)}$}
Consider the problem shown in Section~\ref{sec:problem} and using the notation from Figure~\ref{mlpnet}. 
In the article, when considering the optimization in layer $L^{(new)}$, we have $\ba^{(1)} \in \mathbb{R}^{N \times D_1}$ as the input variable and $\bo^{(new)} \in \mathbb{R}^{N \times D_n}$ as the response variable, 
$\bW^{(new1)} \in \mathbb{R}^{D_1 \times D_n}$ as the weight matrix of layer $L^{(new)}$.
When considering the optimization in layer $L^{(2)}$, we have $\ba^{(new)}\in \mathbb{R}^{N \times D_n}$ as the input variable and  $\bo^{(2)\prime} \in \mathbb{R}^{N \times D_2}$ as the response variable.

\textbf{Algorithm~\ref{algframe}}: 
In Algorithm~\ref{algframe}, we firstly initialize $\mathbf{W}^{(new1)}$ according to the activation function (i.e. we have different initialization methods for different activation functions, e.g. TanH, Sigmoid or ReLU functions) \cite{DBLP:journals/corr/Kumar17,DBLP:journals/corr/HeZR015}. 
And then we get $\bo^{(new)}$ by forward propagation from layer $L^{(new)}$ with the initialized weight $\bW^{(new1)}$, i.e. $\bo^{(new)}$ = $\ba^{(1)} \times \bW^{(new1)}$.
Afterwards we sparsify the neurons in $\bW^{(new1)}$ by Lasso-related algorithms through minimizing Eq~\eqref{lossalg1},
\begin{equation}\label{lossalg1}
\small{
\begin{aligned}
loss_1^{(alg1)}
= \frac{1}{2N}\bsum_{i=1}^{D_n}\left\Vert\mathbf{O}^{(new)}_i-\beta_i \ba^{(1)} \bW^{(new1)}_i\right\Vert_2^2 
+ \lambda (||\bm{\beta}||_1+\frac{\alpha}{2}|\bm{\beta}|^\top \mathbf{R}|\bm{\beta}|),
\end{aligned}
}
\end{equation}
where the similarity matrix $\bR$ is defined by Eq~\eqref{R}.
We can note that
$\bsum_{i=1}^{D_n}||\mathbf{O}^{(new)}_{i}-\ba^{(1)} (\beta_i \bW^{(new1)}_{i})||_2 ^2$ in Eq~\eqref{lossalg1} 
is equal to $||\bo^{(new)} -\ba^{(1)}  (\bW^{(new1)}  \cdot diag(\bbeta)) ||_F^2$, where $diag(\cdot)$ operator maps a vector to a diagonal matrix whose diagonal is the elements of the
vector. 
More detailed analysis of this sparse optimization is delayed to Section~\ref{sec:sparse_neurons}. 

Finally, we optimize $\bW^{(new2)}$ via least squares:
\begin{equation}\label{alg1:least_squares}
loss_2^{(alg1)}= \frac{1}{2N} \left\Vert\mathbf{O}^{(2)\prime}- \ba^{(new)}W^{(new2)}\right\Vert_F^2 = \frac{1}{2N} \left\Vert vec(\mathbf{O}^{(2)\prime})- vec(\ba^{(new)}W^{(new2)})\right\Vert_2^2,
\end{equation}
where $vec(\cdot)$ is the vectorization operator.
Here we can also use ridge regression to penalize the scale of $\bW^{(new2)}$. Note here that we use $\bo^{(2)\prime}$ as the target response because we want our output $\bo^{(2)}=\ba^{(new)}W^{(new2)}$ to be as close to the original output $\bo^{(2)\prime}$ as possible.

\textbf{Algorithm~\ref{altnativeupdating}}: In Algorithm~\ref{algframe}, we only optimize $\bm{\beta}$ by minimizing $loss_1^{(alg1)}$. The other choice is that we optimize  $\bm{\beta}$ and $\bW^{(new1)}$ by minimizing $loss_1^{(alg2)}$ alternatively as shown in Algorithm~\ref{altnativeupdating}. The superscripts of $loss_1^{(alg2)}$ and $loss_1^{(alg1)}$, $loss_2^{(alg2)}$ and $loss_2^{(alg1)}$ are used to distinguish different algorithms but they are exactly the same in Algorithm~\ref{algframe} and~\ref{altnativeupdating}. For optimizing $\bW^{(new2)}$, Algorithm~\ref{altnativeupdating} takes the same operation as Algorithm~\ref{algframe}. The motivation of the second algorithm is to reconstruct the output after sparsifying.


\subsubsection{Reduce the neuron size of $L^{(new)}$ from $\mathbf{W}^{(new2)}$}
\textbf{Algorithm~\ref{pruningbyw2}}:
To reduce the size of $L^{(new)}$ and its corresponding weight matrix $\bW^{(new1)}$, another choice is to make $\bW^{(new1)}$ of size  $D_1 \times D_n$  smaller by reducing $\bW^{(new2)}$ of size $D_n \times D_2$ as shown by Algorithm~\ref{pruningbyw2} because $\bW^{(new2)}$ shares the first dimension with the feature dimension of $\bW^{(new1)}$ which is $D_n$.

In this algorithm, we firstly initialize $\bW^{(new1)}$ according to the activation function same as Algorithm~\ref{algframe}. Then we forward propagate to get $\ba^{(new)}$ = $h(\ba^{(1)} \times \bW^{(new1)})$. Afterwards, we initialize $\bW^{(new2)}$ via least squares as shown in Eq~\eqref{alg1:least_squares} but we denote Eq~\eqref{alg1:least_squares} as $loss_2^{(alg3)}$ to differentiate the algorithms. Finally, we sparsify the neurons in layer $L^{(new)}$ by:
\begin{equation}\label{loss1alg3}
\small{
	\begin{aligned}
loss_1^{(alg3)}&=\frac{1}{2N}\left\Vert \bo^{(2)\prime} -\bsum_{i=1}^{D_n}\beta_i \ba^{(new)}_{i} \bW^{(new2)}_{i,:}\right\Vert_F^2 
+\lambda (\left\Vert\bm{\beta}\right\Vert_1+\frac{\alpha}{2}|\bm{\beta}|^\top \mathbf{R}\left|\bm{\beta}\right|),
\end{aligned}}
\end{equation}
where $\bo^{(2)\prime}$ is the original output of Layer $L^{(2)}$ as shown in Figure~\ref{mlp1}.

However, Algorithm~\ref{pruningbyw2} has much higher complexity compared with Algorithm~\ref{algframe} and~\ref{altnativeupdating} due to the sum operation $\bsum_{i=1}^{D_n}\beta_i \ba_{i}^{(new)} \bW_{i,:}^{(new2)}$. 
And this process takes a longer time to compute than Algorithm~\ref{algframe} and~\ref{altnativeupdating}. 
 Especially, when applying Algorithm~\ref{pruningbyw2} to convolutional layer, the cost time is scaled by the number of channels and the size of matrix is usually very large so that the speed is very slow and sometimes memory error happens. To accelerate the Algorithm~\ref{pruningbyw2}, we can sample on the intermediate output (rows of $\ba^{(new)}_i$) of the layers according to \cite{DBLP:journals/corr/HeZS17}. 


An important computational property for Eq~\eqref{loss1alg3} is that the first term can be rewritten as:
\begin{equation}\label{equation:rewrite_alg3_frobenius_2norm}
\small{
	\begin{aligned}
\frac{1}{2N}\left\Vert vec(\bo^{(2)\prime}) -
\left[vec(\mathbf{T}_1), vec(\mathbf{T}_2), \cdots, vec(\mathbf{T}_{D_n})\right]\bbeta \right\Vert_2^2 ,
\end{aligned}}
\end{equation}
where $\mathbf{T}_i = \ba^{(new)}_{i} \bW^{(new2)}_{i,:}, \forall i \in \{1,2,\cdots, D_n\}$ and $vec(\cdot)$ is the vectorization operator.

\subsection{Sparsify the neurons in the newly added layer}\label{sec:sparse_neurons}
\subsubsection{Interpretation of Lasso in our algorithms}
Consider the sparse optimization in Algorithm~\ref{algframe} and~\ref{altnativeupdating}, let $\bm{\beta}=(\beta_1,\beta_2,...,\beta_i,...,\beta_{D_n})^\top$, Lasso optimization for $\bbeta$ in our problem is defined by
\begin{equation}\label{lassoc}
\scriptsize{
	\bm{\beta} = \argminl_{\bm{\beta}} \frac{1}{2N}\left\Vert\bo^{(new)} -\ba^{(1)}  \bW^{(new1)}  \cdot diag(\bbeta)\right\Vert_F^2+\lambda \left\Vert\bm{\beta}\right\Vert_1, \textit{where}\, \bo^{(new)} = \ba^{(1)}  \bW^{(new1)}.}
\end{equation}
Lasso can give the solution with some exact zeros. So when $\beta_i$ is zero, the corresponding feature $\bW_i^{(new1)}$ will be removed. 

In our method, $\beta_j$ should be in the range of $[0,1]$ to indicate the importance of each neuron.
But in our proposal, we initialized $\bo^{(new)}$ by $\bo^{(new)}= \ba^{(1)} \cdot \bW^{(new1)}$. We can easily find that each $\beta_i$ cannot be negative value when the algorithm converges, because it will cause larger loss in both first term and second term of Eq~\eqref{lassoc} when $\beta_i <0$ than $\beta_i =0$; and also each $\beta_i$ cannot be larger than 1 because it will cause loss in first term of Eq~\eqref{lassoc} and impose larger loss in second term of Eq~\eqref{lassoc} than the loss of the second term when each $\beta_i=1$. So this constraint can be relaxed. 


\subsubsection{Independently interpretable Lasso}
In our work, we use a modification of Lasso algorithm which is called independently interpretable Lasso (iiLasso) \cite{2017arXiv171101796T} that can suppress selecting correlated variables by penalizing the similarity of the predictable variables.
In our problem iiLasso is defined by Eq~\eqref{equation:alg1_iilasso}, 
\begin{equation}\label{equation:alg1_iilasso}
\scriptsize{
\begin{aligned}
\bm{\beta} &= \argminl_{\bm{\beta}} \frac{1}{2N}||\bo^{(new)} - \bX \cdot \mathbf{B}||_F^2
+\lambda (||\bm{\beta}||_1+\frac{\alpha}{2}|\bm{\beta}|^\top \bR |\bm{\beta}|),
\end{aligned}
}
\end{equation}
where $\bX=\bo^{(new)}=\ba^{(1)}  \bW^{(new1)} \in \mathbb{R}^{N\times D_n}$, $\mathbf{B}=diag(\bbeta)$ and $\bR\in \mathbb{R}^{D_n \times D_n}$ is a symmetric matrix whose component $\bR_{jk} \geq 0$ represents the similarity between $\bX_j$ and $\bX_k$ and its component is defined by Eq~\eqref{R}:
\begin{equation}
\scriptsize
\label{R}
\begin{aligned}
& \textbf{standardize } \bo^{(new)}_i, \text{and } \bX_i, \forall i \in \{1, 2, \cdots, D_n\}\\
&\mathbf{r_{jk}} = \frac{1}{N}\mathbf{|X^T_jX_k|},    
\,\, \bR_{jk} =\begin{cases}
\frac{|\mathbf{r}_{jk}|}{1-|\mathbf{r}_{jk}|} &\text{$j \neq k$},\\
0 &\text{$j=k$}.
\end{cases}
\end{aligned}
\end{equation}
The last term of the Eq~\eqref{equation:alg1_iilasso} can also be written as 
$\frac{\lambda \alpha}{2} \sum_{j=1}^{D_n}\sum_{k=1}^{D_n} \bR_{jk}|\beta_j||\beta_k|$.
In this case, if the correlation between two certain neuron variables
becomes higher, i.e. $\mathbf{r}_{jk}\rightarrow 1$, we penalize larger for the two neurons. When $\bR_{jk}$ goes infinity, we can set either $\beta_j$ or $\beta_k$ to be zero. We can easily interpret that when two inputs $\bX_j$ and $\bX_k$ are highly similar, we will impose larger loss in the last term of Eq~\eqref{equation:alg1_iilasso} so that we will enforce either $\beta_j$ or $\beta_k$ to be closer to zero. 
And notice that $\bX=\ba^{(1)}  \bW^{(new1)}$ so that the penalization of two neurons for the similarity is decide both by activation output $\ba^{(1)}$ and weight matrix $\bW^{(new1)}$.
And because all the components in the last term of Eq~\eqref{equation:alg1_iilasso} is not negative, i.e. $|\beta_j| \geq 0$, $\bR_{jk}\geq 0$, we can still relax the constraint that $\beta_j \in [0,1]$. 
And this algorithm can easily be extended to Algorithm~\ref{pruningbyw2} from Eq~\eqref{loss1alg3}.



\savebox{\algleft}{%
\begin{minipage}{.59\textwidth}
\RestyleAlgo{boxruled}
\IncMargin{1.0em}
\begin{algorithm}[H]
	\scriptsize
	\BlankLine
	\Begin{		
		Step 1: Initialize $\mathbf{W}^{(new1)}$ accordingly and initialize $\bm{\beta}$ to ones vector\;
		Step 2: Compute $\bo^{(2)\prime}$, $\ba^{(1)} $, $\mathbf{O}^{(new)}$ and $\bX=\ba^{(1)} \bW^{(new1)}$\;
		Step 3: Standardlize $\bo^{(new)\prime}$ and $\bX$ so that columon vector: $mean(\bo^{(2)\prime}_i)=0$, $mean(\bX_i)=0$ and $\bX_i^\top \bX_i = N$\;
		Step 4: Compute similarity matrix $\bR$ of $\bX$\;
		Step 5: \
		\While{$i < max\_itr$ or $||\bm{\beta}||_0 < c$}{
			$ \bm{\beta}= \argminl_{\bm{\beta}} loss_1^{(alg1)}$\;
		}
		Step 6: Drop the column $i$ of $\mathbf{W}^{(new1)}$ if $\beta_i$ is zero\;

		Step 7: Compute $\ba^{(new)} = h(\ba^{(1)} \mathbf{W}^{(new1)})$\;
		Step 8: $\mathbf{W}^{(new2)} = \argminl_{\mathbf{W}^{(new2)}} loss_2^{(alg1)}$\;
		return $\mathbf{W}^{(new1)}$ and $\mathbf{W}^{(new2)}$.
	}
	\caption{Sparsify $\mathbf{W}^{(new1)}$, optimize $\mathbf{W}^{(new2)}$}
\label{algframe}
\end{algorithm}\DecMargin{6.1em}
\end{minipage}}%
\savebox{\algright}{%
\begin{minipage}{.4\textwidth}
\RestyleAlgo{boxruled}
\IncMargin{0.0em}
\begin{algorithm}[H]
	\footnotesize
\BlankLine
\Begin{		
    Step 1, 2, 3, 4 same as Algorithm~\ref{algframe}\;
	Step 5:
	\While{\
		$i < max\_itr$ or $||\bm{\beta}||_0 < c$}{
		Step 5.1: Fix $\bW^{(new1)}$,
		optimize $ \bm{\beta} = \argminl_{\bm{\beta}} loss_1^{(alg2)}$\;
		Step 5.2: Fix $\bm{\beta}$,
		optimize $ \bW^{(new1)}= \argminl_{ \bW^{(new1)}} loss_1^{(alg2)}$\;
	}
	Step 6, 7, 8 same as Algorithm~\ref{algframe} \;
	return $\mathbf{W}^{(new1)}$ and $\mathbf{W}^{(new2)}$.
}
\caption{Alternatively update}
\label{altnativeupdating}
\end{algorithm}\DecMargin{0.0em}
\end{minipage}}%

\noindent\usebox{\algleft}\hfill\usebox{\algright}%

\RestyleAlgo{boxruled}
\IncMargin{2.0em}
\begin{algorithm}[!ht]
	\footnotesize
	\BlankLine
	\Begin{		
		Step 1: Initialize $\mathbf{W}^{(new1)}$ according to the activation function and initialize $\bm{\beta}$ to ones vector \;
		Step 2: Compute $\bo^{(2)\prime}$, $\ba^{(1)} $, $\mathbf{A}^{(new)}$ and $\mathbf{T}_i$, where
		$\mathbf{T}_i = \ba^{(new)}_{i} \bW^{(new2)}_{i,:}, \forall i \in \{1,2,\cdots, D_n\}$\;
		Step 3: $\mathbf{W}^{(new2)} = \argminl_{\mathbf{W}^{(new2)}} loss_2^{(alg3)}$\;
		Step 4: Standardlize $vec(\bo^{(2)\prime})$ and $vec(\mathbf{T}_i)$ so that $mean(vec(\bo^{(2)\prime}))=0$ and $mean(vec(\mathbf{T}_i)^\top vec(\mathbf{T}_i))=1$\;
		Step 5: Compute similarity matrix $\bR$ from $\mathbf{T}_i$\;
		Step 6:\
		\While{$i < max\_itr$ or $||\bm{\beta}||_0 < c$}{
			$ \bm{\beta} = \argminl_{\bm{\beta}} loss_1^{(alg3)}$\;
		}
		Step 7: Drop column $i$ of $\mathbf{W}^{(new1)}$ and row $i$ of $\mathbf{W}^{(new2)}$ if $\beta_i$ is zero\;
		return $\mathbf{W}^{(new1)}$ and $\mathbf{W}^{(new2)}$.
	}
	\caption{Reduce $\bW^{(new1)}$ by optimizing $\bW^{(new2)}$}\label{pruningbyw2}
\end{algorithm}\DecMargin{0.0em}

\subsection{Optimization solution for $\mathbf{\beta}$}
$\bm{\beta}$ is a vector whose every component corresponds to one feature. We use coordinate descent to tackle this optimization problem. And when $\beta_j = 0$, we drop the corresponding feature (i.e. neuron in neural networks). 
Lemma~\ref{lemma:iilasso_coordinate} gives the closed form solution for the update of each coordinate $\beta_j$ in Algorithm~\ref{algframe},~\ref{altnativeupdating} and~\ref{pruningbyw2}.

\begin{lemma}\label{lemma:iilasso_coordinate}
For the iiLasso problem in Algorithm~\ref{algframe} and~\ref{altnativeupdating}, we can get the best value for each coordinate by closed form: 
\begin{equation}\label{solutionalg1alg2}
\scriptsize{
	\beta_j =\frac{1}{1+\alpha \lambda \bR_{jj}}S\left(\frac{1}{N}\bo^{(new)\top}_{j}\bX_j, \, \lambda(1+\alpha \sum_{c=1,c \neq j}^{D_n}\bR_{jc}|\beta_c|) \right)
},
\end{equation}
And for Algorithm~\ref{pruningbyw2}, the solution is :
\begin{equation}\label{solutionalg3}
\scriptsize{
	\begin{split}
	\beta_j =\frac{1}{1+\alpha \lambda \bR_{jj}}S(\frac{1}{N\times D_2}[vec(\bo^{(2)\prime})-\sum \limits_{i=1,i \neq j}^{D_n} \beta_i \cdot vec(\mathbf{T}_i)]^\top \mathbf{T}_j,  \lambda(1+\alpha \sum_{c=1,c \neq j}^{D_n}\bR_{jc}|\beta_c|)),
\end{split}}
\end{equation}
where 
\begin{equation}
	\label{Rr}
	\scriptsize{
	\begin{aligned}
		&	S(a,b)=sgn(a)(|a|-b)_+=\begin{cases}
			a-b, & \text{$a > 0$ and $b < |a|$}, \\
			a+b, & \text{$a < 0$ and $b< |a|$ }, \\
			0 ,	   & \text{$|a|<b$}.\end{cases}
	\end{aligned}}
\end{equation}
\end{lemma}

\subsection{Optimization solution for $\bW^{(new1)}$}
In the Algorithm~\ref{altnativeupdating}, for min $loss_1^{(alg2)}$, we can get the solution directly by $\frac{\partial loss_1^{(alg2)}}{\partial \bW^{(new1)}}=0$. And we get $\bW^{(new1)} = (\bc^\top\bc)^{-1}\bc^\top\bo^{(new)}$, where $\bc =\ba^{(1)} \cdot diag(\bm{\beta}) $. 

\subsection{Optimization solution for $\bW^{(new2)}$}
The optimization of $\bW^{(new2)}$ for $loss_2^{(alg1)}$ and  $loss_2^{(alg2)}$ in Algorithm~\ref{algframe} and~\ref{altnativeupdating} is a problem of least squares. 
We take the gradient of $loss_2^{(alg1)}$ or $loss_2^{(alg2)}$ w.r.t. $\bW^{(new2)}$ and set this gradient to be 0: 
\begin{equation}\label{nw2j}
\scriptsize
	\bW^{(new2)} = (\ba^{(new) \top} \ba^{(new)})^{-1}\ba^{(new) \top} \bo^{(2)\prime}.
\end{equation}

\section{Experiment} \label{experiment}
We empirically compare CompNet with NetMorph \cite{wei2016network} on MNIST and CIFAR10 dataset. We apply our algorithms in three neural network structures, namely LeNet, VGG16 (VGG D with 16 layers) and VGG19 (VGG E with 19 layers) \cite{lecun2015lenet,SimonyanZ14a} respectively. 
LeNet with 4 layers is denoted as LeNet4. 

In all experiments, we morph the well trained neural network with an additional layer which contains potentially redundant neurons or channels. We denote the neuron size before the sparse optimization as $N\_redundant$.
Then, we use the same neuron size $N\_redundant$ in NetMorph to compare, termed as \textit{NetMorph-Redundant}. And in addition, we equip NetMorph with an unfair advantage to manually set the number of neurons to be $N\_sparse$ which is the neuron size after sparse optimization in Algorithm~\ref{altnativeupdating} to compare. We term this NetMorph setting to be \textit{NetMorph-Oracle}.


\subsection{Parameter setting up}
For MNIST, we morph from LeNet4 to LetNet5 by adding a convolutional layer. MNIST of handwritten digits includes a training set of $60,000$ examples, and a test set of $10,000$ examples that are used as training dataset and validation dataset respectively. We use $5e{-3}$ and $1e{-6}$ as learning rate and weight decay respectively in LeNet experiments. 
For CIFAR10, we morph from VGG15 to VGG16, from VGG16 to VGG17 and from VGG18 to VGG19 by adding one convolutional layer respectively. It is popular to insert batch normalization layers to VGG models. However, we don't use batch normalization for simplicity and it can be easily extended to the batch normalization setting for our algorithms. We use $3e{-4}$ and $1e{-6}$ as the learning rate and weight decay respectively in VGG experiments. 
CIFAR10 contains $50, 000$ training images and $10, 000$ test images that are used as training dataset and validation dataset respectively.
We also use some common techniques to prevent over-fitting such as dropout, $l_2$ regularizer and data augmentation. 
For the optimization method, we use SGD with momentum value of $0.9$. And we set the $
\lambda$ and $\alpha$ in iiLasso to be 0.1 to test.

\begin{wraptable}{r}{0.8\textwidth}
\centering
\scriptsize
{
\begin{tabular}{|c|c|c|c|c|c|}
\hline
	\,	& N\_redundant & Alg1                      & Alg2                      & Alg3 & Avg compress rate  \\ \hline
		LeNet4 to LeNet5 (ReLU) & 100      & 41                         & 46  & 49    & 45.3\%               \\ \hline
		VGG15 to VGG16 (ReLU) & 512      & 165                        & 200 & 255   & 40.4\%                 \\ \hline
		VGG15 to VGG16 (Sigmoid) & 512      & 196         & 162 & 105   & 30.1\%                 \\ \hline
		VGG15 to VGG16 (TanH) & 512   & 213      & 178                      & 157   & 35.7\%                 \\ \hline
		VGG16 to VGG17 (ReLU) & 256      & 155 & 121 & 127   & 52.5\%               \\ \hline
		VGG18 to VGG19  (ReLU) & 512      & 221                        & 211 & 255   & 44.5\%               \\ \hline
\end{tabular}}
\caption{Number of neurons before and after sparsifying for each algorithm.}
\label{tablencompressionratio}
\end{wraptable} 
\subsection{LeNet4 to LeNet5}\label{lenet}
\begin{wrapfigure}{L}{0.71\textwidth}
	\centering
	\subfloat[LetNet4 to LeNet5]{%
		\includegraphics[width=0.35\textwidth]{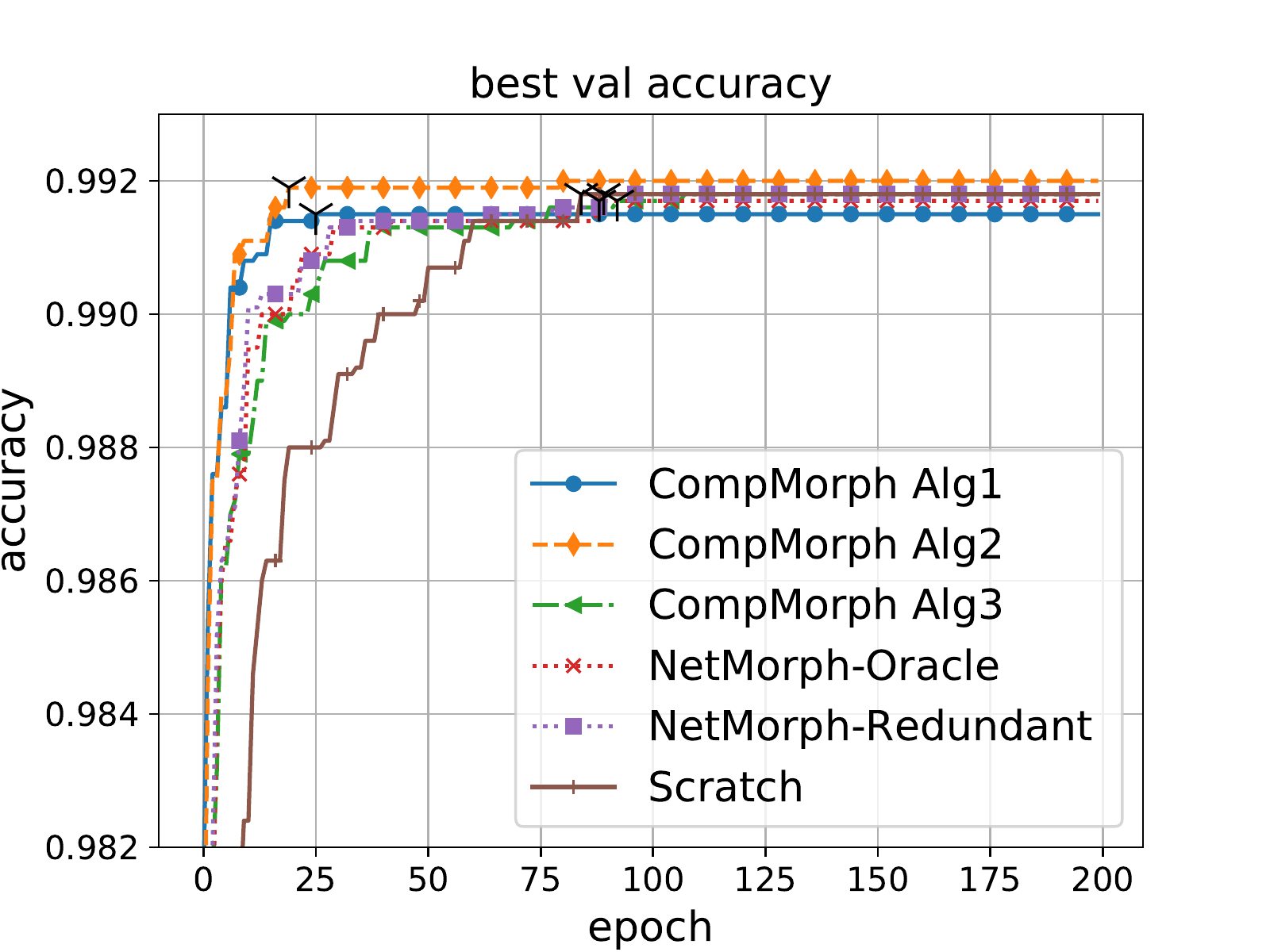}
		\label{mnistvalacc}
	} 
	\subfloat[VGG15 to VGG16]{%
		\includegraphics[width=0.35\textwidth]{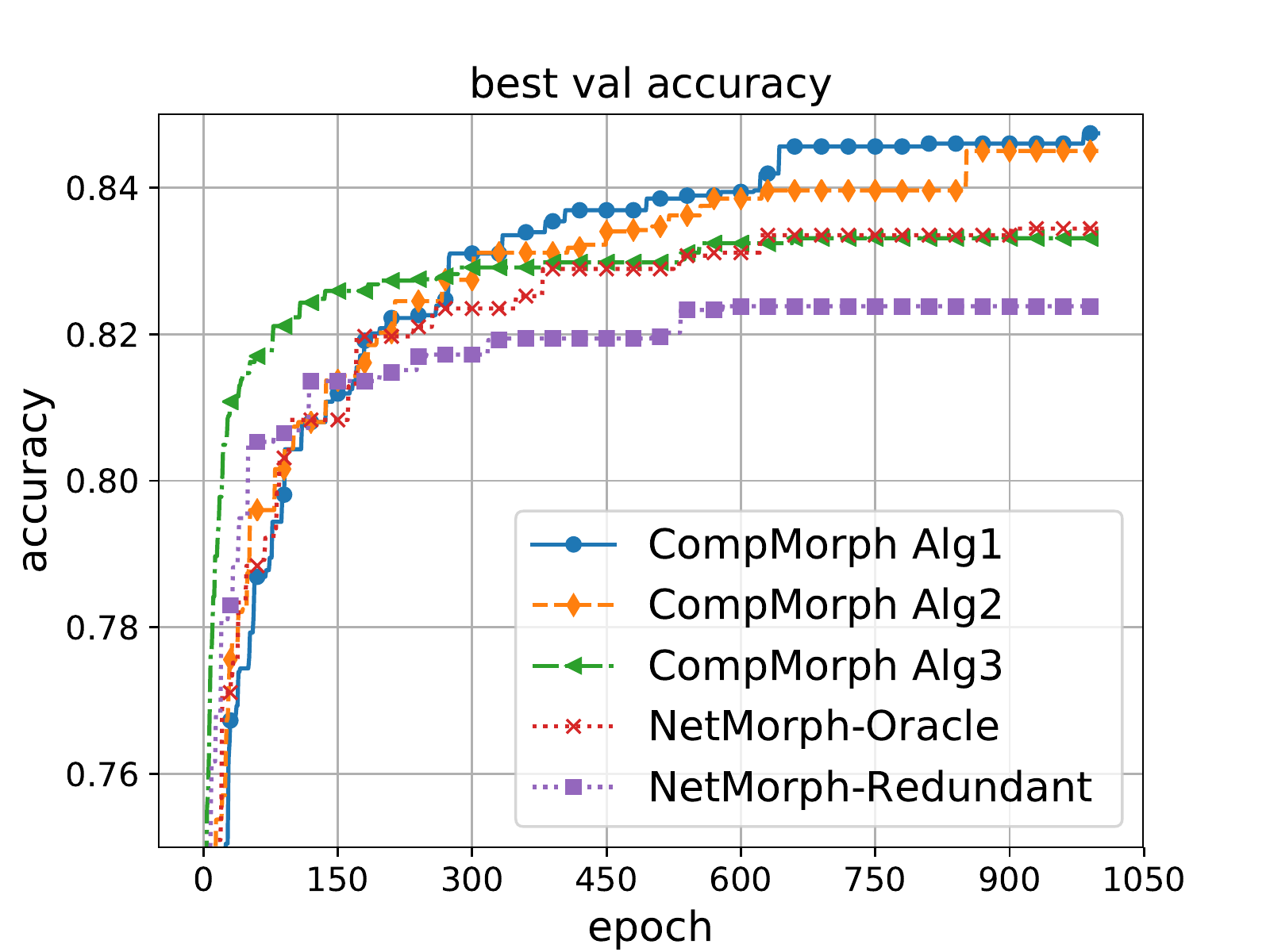}
		\label{reludecay1e-6best}
	} 
	\caption{Left: LetNet4 to LeNet5 with ReLU activation function. Right: VGG15 to VGG16 with ReLU activation function.}
	\label{mnistlenet_and_1516}
\end{wrapfigure}

We train LeNet4 for 200 epochs on MNIST dataset as the parent neural network to morph. And then we insert one convolutional layer with 100 filters and ReLU activation function between the two convolutional layers of LeNet4, denoted as LetNet5. 
The result is shown in Figure~\ref{mnistvalacc} and Table~\ref{valacctable}.
We can see that 
CompNet Alg1 gets 99.15\% accuracy on epoch 25, 
CompNet Alg2 gets 99.19\% accuracy on epoch 19, 
CompNet Alg3 gets 99.17\% accuracy on epoch 92, 
NetMorph-Oracle gets 99.17\% on epoch 88
and NetMorph-Redundant gets 99.18\% on epoch 89 and
Scratch gets 99.19\% on epoch 84. Our algorithms and NetMorph can converge fast. But in this case, CompNet Alg2 converges fastest and gets best accuracy rate. And Table~\ref{tablencompressionratio} shows the $N\_sparse$ and the average compression ratio for our three algorithms. 

%

\begin{table}[h!]
\centering
\footnotesize
{
\begin{tabular}{|c|c|c|c|l|l|l|}
\hline
		& Alg1           & Alg2           & Alg3           & \begin{tabular}[c]{@{}l@{}}NetMorph-\\ Oracle\end{tabular} & \begin{tabular}[c]{@{}l@{}}NetMorph-\\ Redundant\end{tabular}  & Scratch\\ \hline
		LeNet4 to LeNet5 (ReLU) & 99.15\%          & \textbf{99.20\%} & 99.18\%          & 99.17\%          & 99.18\%      &    99.19\%    \\ \hline
		VGG15 to VGG16 (ReLU)    & \textbf{84.74\%}          & 84.50\% & 83.31\%          & 83.44 \%          & 82.38\%       &   \textbf{83.14}\%   \\ \hline
			VGG15 to VGG16 (Sigmoid)    & 83.63\%          & \textbf{83.71\%} & 83.08\%          & 83.25\%          & 82.54\%      &      69.97\% \\ \hline
				VGG15 to VGG16 (TanH)    & 83.68\%         & 83.52\% & \textbf{83.86\%}        & 83.41\%          & 83.00\%      &    80.70\%   \\ \hline
		VGG16 to VGG17 (ReLU)    & 85.45\%          & 85.37\%          & \textbf{85.78\%} & 85.69\%          & 85.51\%      &   \textbf{82.71}\%    \\ \hline
		VGG18 to VGG19 (ReLU)    & \textbf{89.00\%} & 88.84\%          & 88.90\%          & 88.95\%          & 88.87\%        &  85.26\%   \\ \hline
\end{tabular}
}
\caption{Best validation accuracy in different experiments.}	\label{valacctable}
\end{table}


\subsection{VGG15 to VGG16} \label{vgg15to16}
We firstly remove the first convolutioanl layer with $512$ filters and ReLU activation function from VGG16, denoted as VGG15. 
Afterwards, we traine VGG15 from scratch for enough long time, i.e., 1,000 epochs, until its validation accuracy converges. Then we insert the removed convolutioanl layer back with the initial $512$ filters and ReLU activation function to VGG15. 
The results are shown in Figure~\ref{reludecay1e-6best}.
It can be seen that Algorithm~\ref{pruningbyw2} converges fastest, and Algorithm~\ref{algframe} and~\ref{altnativeupdating} are much better than NetMorph-Redundant and NetMorph-Oracle. We also notice that NetMorph-Oracle converges similar to NetMorph-Redundant, but NetMorph-Oracle gets better accuracy which means that the \textit{redundant neurons or channels sparsified by CompNet is really working}. Table~\ref{valacctable} shows the best validation accuracy for different algorithms in this setting and Algorithm~\ref{algframe} works best. 
Finally, we also traine the new model VGG16 from scratch for 2,000 epochs. And best validation accuracy for this case is about $83.14\%$ which is worse than our algorithms that are better than $84.00\%$ and only trained for 1,000 epochs in the same setting. We do not draw the validation curve for scratch in this experiment and the following experiments, because the scratch method converges very slow and the scale of the figure usually ranges from 40\% to 90\% in the vertical axis so that we cannot see the other curves clearly.



\begin{wrapfigure}{!ht}{0.71\textwidth}
\centering
\subfloat[New layer with Sigmoid]{%
	\includegraphics[width=0.35\textwidth]{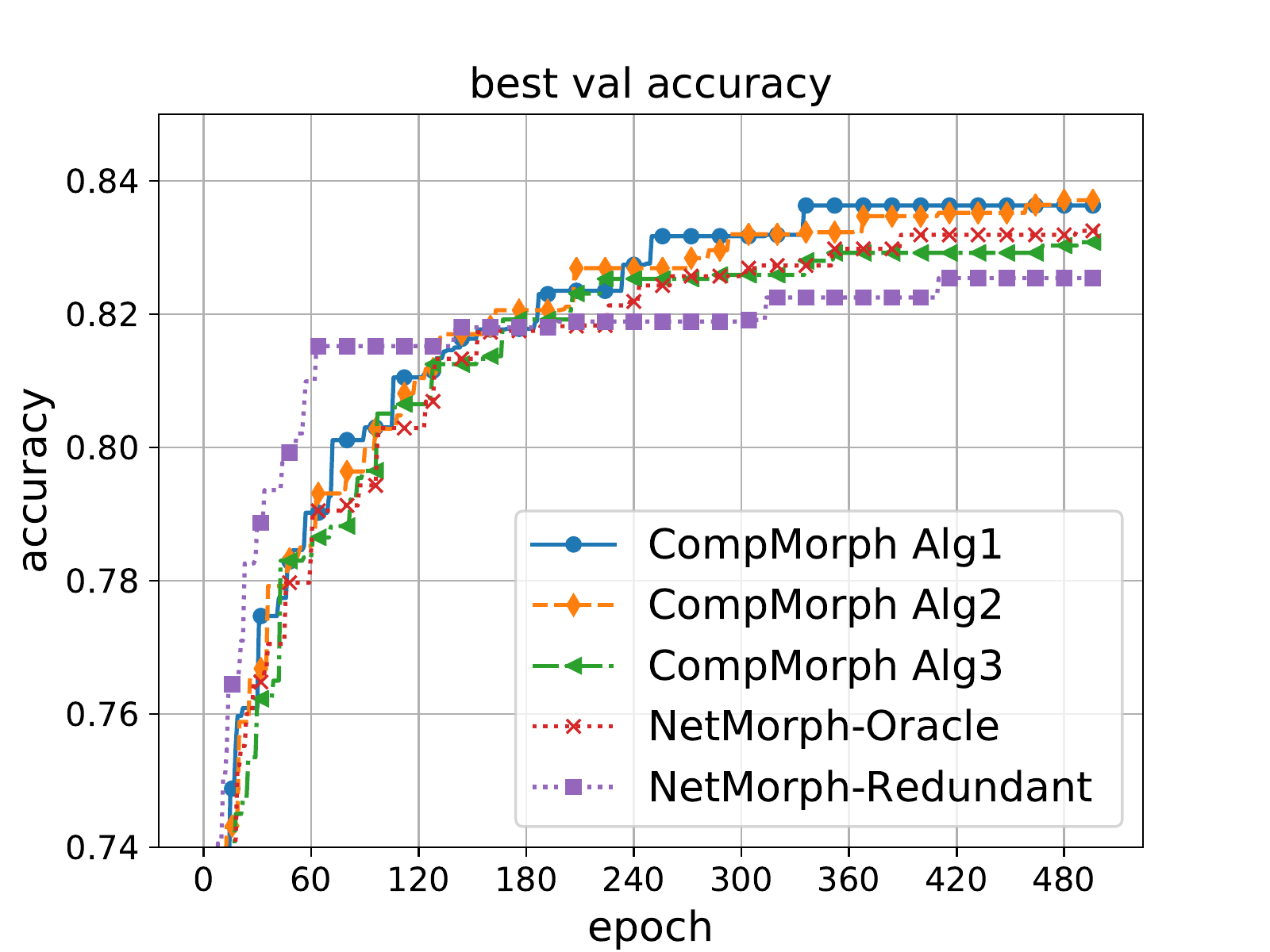}
	\label{15to16sigmoidbest}
} 
\subfloat[New layer with TanH]{%
	\includegraphics[width=0.35\textwidth]{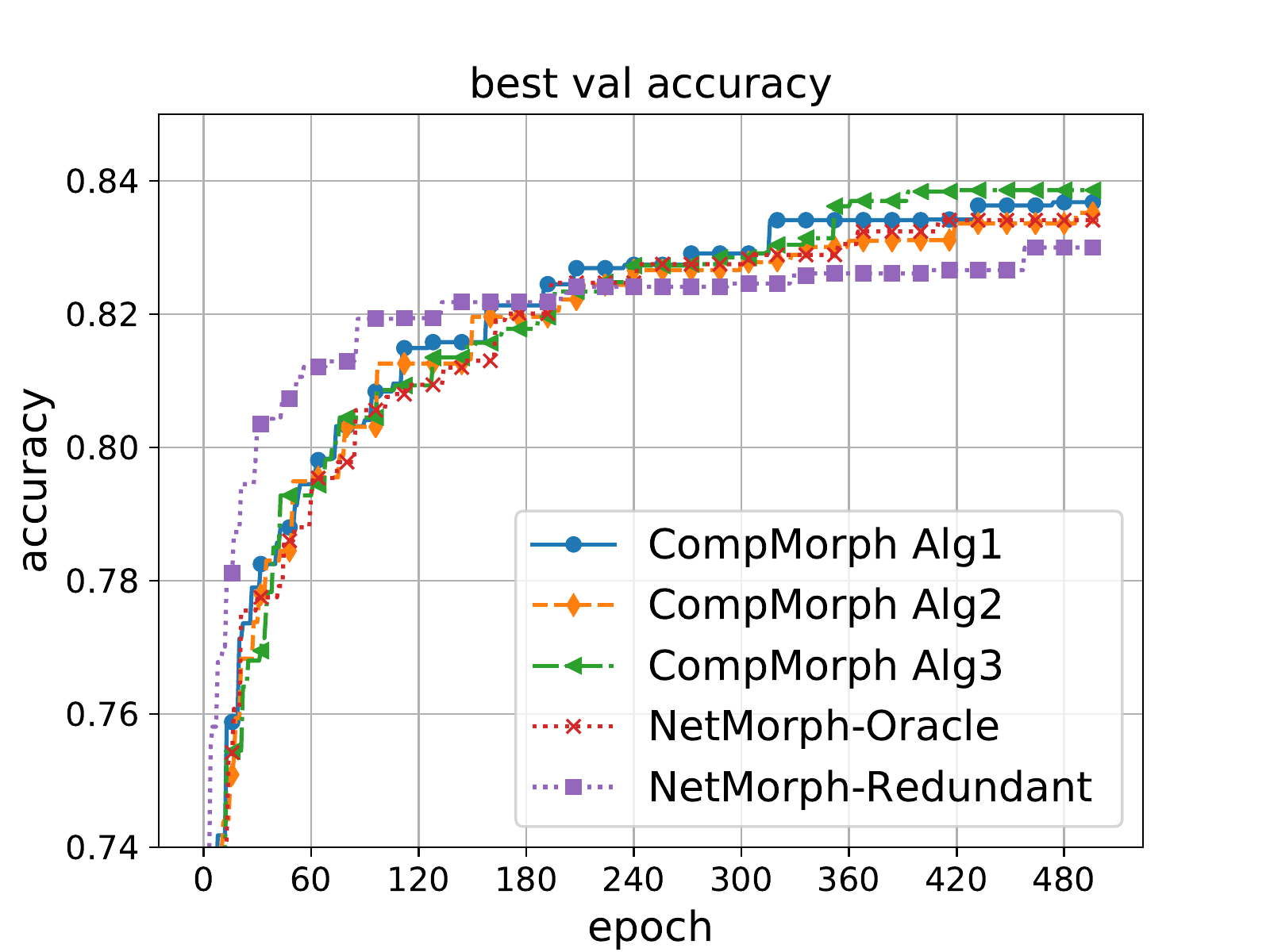}
	\label{15to1tanhbest}
}
\caption{Adding one new layer with Sigmoid and TanH activation function respectively from VGG15 to VGG16.}
\label{sigmoidtanh}
\end{wrapfigure}

Moreover, we also insert a convolutional layer with Sigmoid and TanH activation function and the results are shown in Figure~\ref{sigmoidtanh}. 
In both cases, we can see that Algorithm~\ref{algframe} and~\ref{altnativeupdating} get better validation accuracy than NetMorph-Redundant and NetMorph-Oracle. 
Meanwhile, the validation accuracies training from scratch for TanH and Sigmoid networks with same epochs are $80.70\%$ and $69.97\%$ respectively, which are both worse than CompNet and NetMorph especially for the Sigmoid network. This is potentially because  
Sigmoid with value in range [0,1] is easier to encounter the saturation problem than TanH with value in range [-1,1].


\begin{wrapfigure}{L}{0.71\textwidth}
	\centering
\subfloat[\scriptsize{VGG16 to VGG17 with ReLU function.}]{%
	\includegraphics[width=0.35\textwidth]{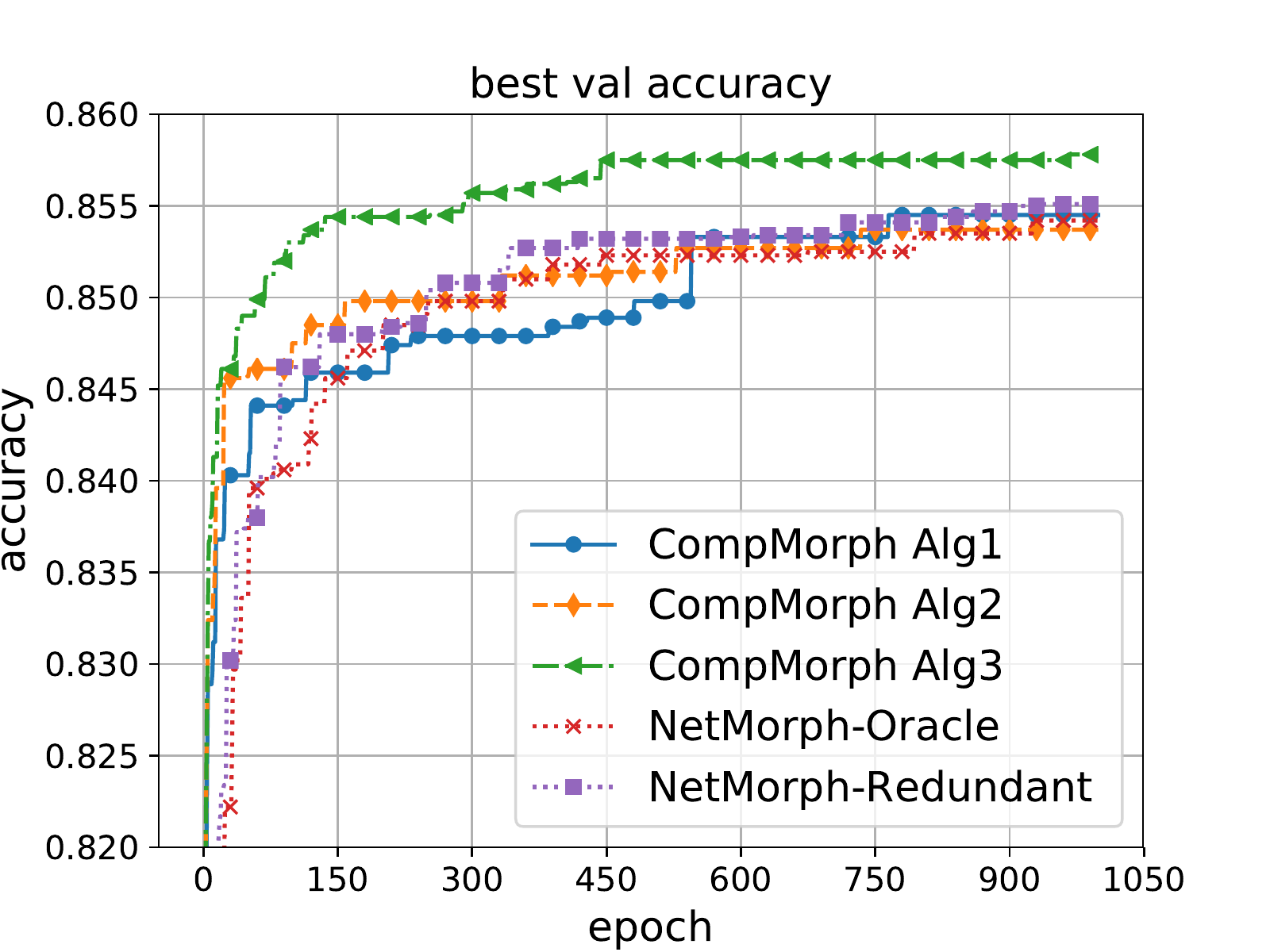}
	\label{17reludecay1e-6best}}
\subfloat[\scriptsize{VGG18 to VGG19 with ReLU function.}]{%
	\includegraphics[width=0.35\textwidth]{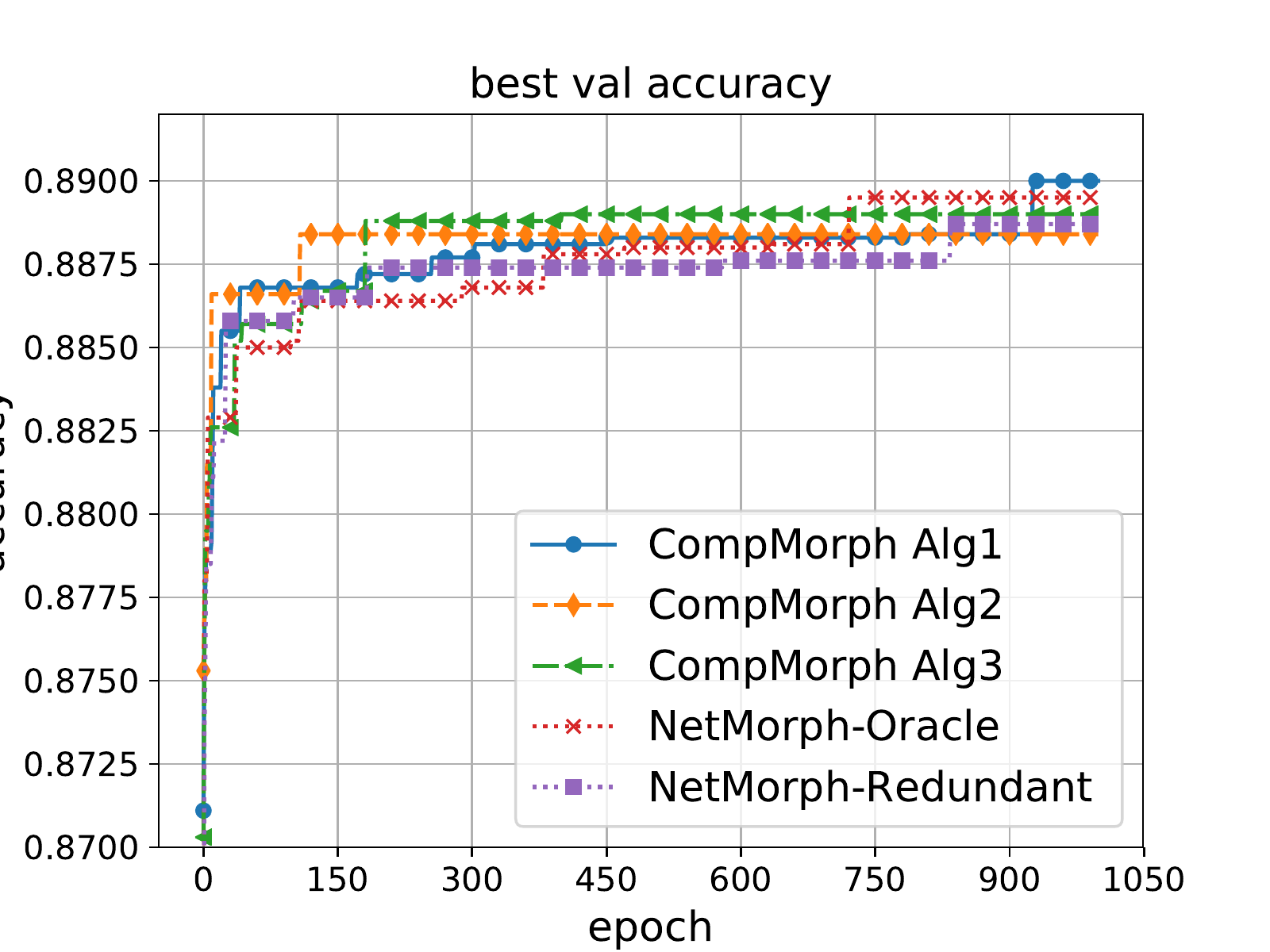}
	\label{1819dc1e-6best}}
	\caption{Left: VGG16 to VGG17 with ReLU activation function. Right: VGG18 to VGG19 with ReLU activation function.}
	\label{1617dir_and_1819}
\end{wrapfigure}

\subsection{VGG16 to VGG17}\label{vgg16tovgg17}
We choose the well tuned model of VGG16 with ReLU activation function from the Section~\ref{vgg15to16} and continue to insert one convolutional layer with 256 filters and ReLU activation to the selected model, denoted as VGG17. 
We also traine VGG17 model from scratch for $2,000$ epochs and its best accuracy is $\textbf{82.71}\%$ which is much lower than the best accuracy $\textbf{83.14}\%$ of VGG16 trained from scratch.
It may be caused by vanishing gradient \cite{DBLP:journals/corr/HeZRS15} and singularities \cite{DBLP:journals/corr/Orhan17}. 
Our algorithms and NetMorph seem to alleviate the two problems according to our results where the best accuracies are higher than $85\%$ in the same setting as shown in Figure~\ref{17reludecay1e-6best}. 
Algorithm~\ref{algframe} and~\ref{altnativeupdating} get similar result as NetMorph-Redundant and NetMorph-Oracle but Algorithm~\ref{pruningbyw2} gets best accuracy. 
Table~\ref{valacctable} summarizes the best accuracy for different algorithms. 

\subsection{VGG18 to VGG19}
Similarly, we firstly remove one convolutional layer with 512 filters and ReLU activation function from VGG19 denoted as VGG18. We train VGG18 until convergence and then add the removed convolutional layer back with our algorithms. 
Figure~\ref{1819dc1e-6best} shows the performance of different algorithms. 
We can see that all the algorithms converge similarly but Algorithm~\ref{algframe} gets best accuracy. The best validation accuracies for different algorithms is summarized in Table~\ref{valacctable}. We also train VGG19 from scratch for 2,000 epochs and it only gets the best accuracy of $85.26\%$ which is much worse than the accuracy of our algorithms. 
\bibliographystyle{epflbibstyle}
{\small
\bibliography{bibcompnet}
}

\end{document}